\pdfoutput=1
\documentclass[11pt]{article}
\usepackage{EMNLP2022}
\usepackage{times}
\usepackage{latexsym}
\usepackage{graphicx}
\graphicspath{ {./images/} }
\newcommand{\keywords}[1]{\textbf{\textit{Keywords---}} #1} 

\usepackage[T1]{fontenc}

\usepackage[utf8]{inputenc}

\usepackage{microtype}

\usepackage{inconsolata}
\usepackage[english]{babel}
\usepackage{blindtext}
\usepackage[table]{colortbl}

\title{Supervised Acoustic Embeddings And Their Transferability Across Languages}

\author{Sreepratha Ram \\
   UAE University\\
  \texttt{sree\_ram@uaeu.ac.ae}  \\
  \And
  Hanan Aldarmaki \\
  MBZUAI \\
  \texttt{hanan.aldarmaki@mbzuai.ac.ae} \\}

\begin{document}
\maketitle
\begin{abstract}
In speech recognition, it is essential to model the phonetic content of the input signal while discarding irrelevant factors such as speaker variations and noise, which is challenging in low-resource settings. Self-supervised pre-training has been proposed as a way to improve both supervised and unsupervised speech recognition, including frame-level feature representations and Acoustic Word Embeddings (AWE) for variable-length segments. However, self-supervised models alone cannot learn perfect separation of the linguistic content as they are trained to optimize indirect objectives. In this work, we experiment with different pre-trained self-supervised features as input to AWE models and show that they work best within a supervised framework. Models trained on English can be transferred to other languages with no adaptation and outperform self-supervised models trained solely on the target languages. 

\end{abstract}


\section{Introduction}
With supervised speech recognition systems getting more robust and accurate due to the availability of large amounts of labeled data and computational power \cite{gulati2020conformer,baevski2020wav2vec}, more attention is now given to low-resource languages for which training data are limited or non-existent \cite{aldarmaki2022unsupervised}. Unsupervised pre-training using unlabeled speech can be leveraged to improve both supervised and unsupervised models; for instance, speech representations pre-trained on large amounts of unlabeled speech from multiple languages have been shown to improve ASR performance for low-resource languages \cite{kawakami2020learning, conneau2020unsupervised}.

While most supervised ASR models operate at the level of phones, word-level segmental ASR where variable-length segments are modeled and embedded into fixed-dimensional vectors have also been explored with relative success \cite{abdel2013deep,he2015segmental}. 
In a similar vein, Acoustic Word Embeddings (AWEs) have been proposed as a way to efficiently compare variable-length speech segments in low-resource settings \cite{peng2020correspondence,kamper2020multilingual}. Unlike written words, spoken words naturally contain speaker and phonetic variability that makes them more difficult to model in a latent space without supervision. Self-supervised pre-training and cross-lingual transfer are two possible approaches to make unsupervised models more robust to non-linguistic variations in the input signal. 

In this work, we investigate the performance of self-supervised training of AWE models versus supervised training with zero-shot cross-lingual transfer. We experiment with different types of acoustic features and measure their performance separately and within the AWE models. While we find that pre-trained acoustic features improve the performance of self-supervised AWE models to some extent, a larger improvement can be achieved when the AWE models are trained in a supervised manner using small amount of labeled data from a different language. This zero-shot cross-lingual transfer is observed consistently across different languages, and particularly with the use of pre-trained feature representations. Our results suggest that supervised training with zero-shot cross-lingual transfer is a more effective approach for low-resource speech models compared with purely self-supervised training\footnote{We provide python training and evaluation scripts for replicating our experiments: \href{https://github.com/h-aldarmaki/acoustic\_embeddings}{https://github.com/h-aldarmaki/acoustic\_embeddings} }.   

\section{Background \& Related Work}
Spoken language is often modeled using short fixed-length frames of 10 to 30 ms duration, which results in variable-length word segments. Dynamic Time Warping (DTW) is an early technique that uses dynamic programming to compare variable-length segments by finding optimal frame-wise alignment. DTW is rather inefficient, which motivates embedding variable-length segments into vectors of fixed size that can be compared using more efficient metrics such as cosine or Euclidean distance \cite{levin2013fixed}. Different types of Acoustic Word Embeddings (AWE) have been proposed. As these techniques are generally meant for low-resource languages, they are typically trained in a self-supervised manner, most commonly using an auto-encoder network with reconstruction loss \cite{chung2016audio, holzenberger2018learning}. Compared with direct comparison via DTW, these AWEs generally result in similar or slightly superior performance while being far more efficient \cite{holzenberger2018learning}.   \citet{peng2020correspondence} describes an alternative training strategy using correspondence auto-encoders, which relies on word pairs extracted via unsupervised spoken term discovery, and further improvements can be achieved using contrastive learning and multi-lingual adaptation \cite{jacobs2021acoustic}. 

The above models use static acoustic features (e.g. MFCCs) as input. \citet{van2021comparison} shows that using pre-trained features like CPC \cite{van2018representation} improves the performance of unsupervised AWE models. Pre-trained features have been repeatedly shown to improve performance in supervised downstream tasks \cite{yang2021superb}. In addition, pre-trained features have been shown to transfer across languages. For instance, a modified version of CPC (MCPC) is described in \citet{riviere2020unsupervised}, which demonstrates that pre-training these features on Egnlish results in improved phone classification accuracy for other languages. Other types of pre-trained features, such as wav2vec 2.0 \cite{Baevski2020wav2vec2A} have been shown to improve both supervised and unsupervised ASR performance \cite{baevski2021unsupervised}, and multi-lingual training of these features (i.e. XLSR-53)  can lead to improvements across many languages compared to monolingual pre-training \cite{conneau2020unsupervised}.

\section{Objectives \& Methodology}
The objective of this study is to investigate the effectiveness and trasnsferability of pre-trained acoustic features when used as input to acoustic word embeddings. To that end, we compare self-supervised AWEs trained directly on the target languages versus zero-shot cross-lingual  transfer of supervised AWEs trained on a different source language. To our knowledge, the combination of pre-trained features with AWE models has not been fully investigated; most AWE models are trained with standard acoustic features like MFCCs, while self-supervised features are typically evaluated within supervised models fine-tuned for the target languages. Furthermore,  zero-shot cross-lingual transfer of supervised AWEs has not been the focus of previous works in this area, which mainly focused on improving self-supervised AWEs.  

For the purpose of this evaluation, we use a relatively simple architecture for the embedding model and we fix the hyper-parameters based on preliminary validation results for English self-supervised AWEs\footnote{We observed that self-supervised models were very sensitive to the choice of architecture and hyper-parameters, so we fixed these in favor of self-supervised models. As shown in later sections, we still got better results with the supervised models, which shows that they are more robust and easier to optimize on top of being more effective. }. We do not do any further tuning of the self-supervised or the supervised models.   
We use English as the source language, and evaluate zero-shot transfer on four other languages: French, German, Spanish, and Arabic, with the latter used as a challenge set since it contains more variability and noise. No labeled data were used for the target languages with the exception of word boundaries which were obtained via force alignment. We evaluate mainly using minimal-pair ABX error rates to measure phonetic discriminability and speaker invariance. We also cluster the embedded words and measure how often different occurrences of the same words end up in the same cluster.

\section{Experimental Settings}

\subsection{Model Architecture}\label{sec:model}
Our AWE model consists of a multi-layer bidirectional LSTM encoder, followed by a uni-directional LSTM decoder, similar to  \citet{chung2016audio} and \cite{holzenberger2018learning}.
The encoder takes a sequence of $T$ acoustic features representing one spoken word. The forward and backward states of the last hidden layer of the encoder are concatenated and used as an embedding of the given word, call it \textbf{h}$^T$. The decoder generates the target sequence one step at a time, conditioned on \textbf{h}$^T$ and the  output at the previous time step, similar to \citet{chung2018speech2vec}. In the self-supervised setting, the target sequence is the same as the input sequence, so the model is trained as an auto-encoder with MSE loss. In the supervised setting, the target is a sequence of phonemes representing the input word, and the model is trained by minimizing the negative log-likelihood. 
We used 2-layer networks with 100 hidden units for most models, which results in embeddings of size 200. 
We also used dropout with probability 0.3 on the input features, similar to the denoising networks used in \citet{chung2016audio}. More details of the parameters and training process can be found in the Appendix. 

\subsection{Feature Extraction}

For easier reproduciblity, we used the s3prl toolkit\footnote{https://github.com/s3prl/s3prl} for extracting all features. We used the pre-trained s3prl upstream models; among the many pretrained self supervised speech representations available, modified CPC, Wav2Wec2 and XLSR-53 were chosen based on superior DTW-based ABX scores\footnote{We did experiment with other features like APC, VQ-APC, and VQ-Wav2Vec, and got similar or inferior performance to MCPC and Wav2Vec2. We opted to omit these for brevity.}. All pre-trained models, with the exception of XLSR, have been exclusively pre-trained on English data. XLSR-53 was pre-trained on unlabeled speech from 53 languages, including all target languages in our experiments. As observed by other researchers \cite{bartelds2022neural}, the performance of features extracted from transformer-based models is largely dependent on the choice of layer; we  used the last hidden layer for modified CPC, the second to last hidden layer for Wav2Vec2 and the central hidden layer (layer 12) for XLSR-53. Averaging all layers gave reasonable results, but these choices led to the best performance. For MFCC features, we also used the s3prl implementation, which includes 13 static features as well as dynamic delta and delta-delta coefficients. 

\subsection{Data}
We used the Librispeech \cite{panayotov2015librispeech} and Multilingual Librispeech \cite{pratap2020mls} datasets for English (en), French (fr), German (de), and Spanish (es). We used the dev sets for training, and test sets for evaluation (dev-clean and test-clean for English). We obtained the word boundaries automatically by forced alignment. For Arabic (ar), we used the dev and test sets of MGB2 \cite{ali2016mgb}. This dataset is expected to be more challenging as it contains a diversity of dialects as well as various noise conditions. See the Appendix for more details on the datasets and the word alignment process. 

\if{false}
\begin{table}
\centering
\begin{tabular}{|l|l|l|}
\hline
Features & Size & Layer \\ \hline
MFCC     & 39   & last    \\ \hline
wav2vec  & 768  & layer 11    \\ \hline
MCPC     & 512  & last     \\ \hline
XLSR-53  & 1024 & layer 12     \\ \hline
\end{tabular}
\captionof{table}{Overview - Features}
\end{table}
\fi

\begin{table*}

\centering
\begin{tabular}{|cccccccccc|}
\cline{2-10}
\multicolumn{1}{c|}{}   & 
  \multicolumn{2}{c|}{en} &
  \multicolumn{2}{c|}{fr} &
  \multicolumn{2}{c|}{de} &
  \multicolumn{2}{c|}{es} &
  ar \\ 
\cline{2-10}
\multicolumn{1}{c|}{} &
  \multicolumn{1}{c|}{within} &
  \multicolumn{1}{c|}{across} &
  \multicolumn{1}{c|}{within} &
  \multicolumn{1}{c|}{across} &
  \multicolumn{1}{c|}{within} &
  \multicolumn{1}{c|}{across} &
  \multicolumn{1}{c|}{within} &
  \multicolumn{1}{c|}{across} &
  across \\  \hline
\cline{2-10}
\multicolumn{10}{|l|}{\cellcolor{black!10}\textit{Using DTW}}                                                                                     \\ \hline
\multicolumn{1}{|l|}{MFCC} &
  \multicolumn{1}{c|}{9.98} &
  \multicolumn{1}{c|}{19.85} &
  \multicolumn{1}{c|}{12.59} &
  \multicolumn{1}{c|}{24.82} &
  \multicolumn{1}{c|}{11.46} &
  \multicolumn{1}{c|}{25.03} &
  \multicolumn{1}{c|}{11.83} &
  \multicolumn{1}{c|}{25.27} &
  40.98 \\ \hline
\multicolumn{1}{|l|}{wav2vec} &
  \multicolumn{1}{c|}{8.51} &
  \multicolumn{1}{c|}{11.15} &
  \multicolumn{1}{c|}{9.95} &
  \multicolumn{1}{c|}{15.61} &
  \multicolumn{1}{c|}{9.01} &
  \multicolumn{1}{c|}{15.08} &
  \multicolumn{1}{c|}{10.13} &
  \multicolumn{1}{c|}{15.46} &
  37.42 \\ \hline
\multicolumn{1}{|l|}{MCPC} &
  \multicolumn{1}{c|}{7.80} &
  \multicolumn{1}{c|}{11.74} &
  \multicolumn{1}{c|}{9.96} &
  \multicolumn{1}{c|}{17.09} &
  \multicolumn{1}{c|}{9.22} &
  \multicolumn{1}{c|}{15.85} &
  \multicolumn{1}{c|}{11.14} &
  \multicolumn{1}{c|}{18.03} &
  38.99 \\ \hline
\multicolumn{1}{|l|}{XLSR-53} &
  \multicolumn{1}{c|}{9.45} &
  \multicolumn{1}{c|}{13.72} &
  \multicolumn{1}{c|}{10.97} &
  \multicolumn{1}{c|}{16.77} &
  \multicolumn{1}{c|}{10.89} &
  \multicolumn{1}{c|}{15.76} &
  \multicolumn{1}{c|}{14.31} &
  \multicolumn{1}{c|}{19.31} &
  40.15 \\ \hline
\multicolumn{10}{|l|}{\cellcolor{black!10}\textit{Self-Supervised AWEs in each language}}                     \\ \hline
\multicolumn{1}{|l|}{MFCC} &
  \multicolumn{1}{c|}{12.30} &
  \multicolumn{1}{c|}{19.12} &
  \multicolumn{1}{c|}{16.63} &
  \multicolumn{1}{c|}{25.06} &
  \multicolumn{1}{c|}{16.71} &
  \multicolumn{1}{c|}{25.99} &
  \multicolumn{1}{c|}{16.58} &
  \multicolumn{1}{c|}{25.21} &
  \multicolumn{1}{c|}{43.92} \\ \hline
\multicolumn{1}{|l|}{wav2vec} &
  \multicolumn{1}{c|}{6.63} &
  \multicolumn{1}{c|}{9.27} &
  \multicolumn{1}{c|}{9.94} &
  \multicolumn{1}{c|}{13.19} &
  \multicolumn{1}{c|}{10.56} &
  \multicolumn{1}{c|}{15.09} &
  \multicolumn{1}{c|}{12.25} &
  \multicolumn{1}{c|}{15.23} &
  \multicolumn{1}{c|}{38.16} \\ \hline
\multicolumn{1}{|l|}{MCPC} &
  \multicolumn{1}{c|}{7.66} &
  \multicolumn{1}{c|}{9.53} &
  \multicolumn{1}{c|}{11.64} &
  \multicolumn{1}{c|}{16.19} &
  \multicolumn{1}{c|}{10.24} &
  \multicolumn{1}{c|}{15.30} &
  \multicolumn{1}{c|}{13.25} &
  \multicolumn{1}{c|}{16.29} &
  \multicolumn{1}{c|}{41.09} \\ \hline
\multicolumn{1}{|l|}{XLSR-53} &
  \multicolumn{1}{c|}{10.61} &
  \multicolumn{1}{c|}{12.19} &
  \multicolumn{1}{c|}{13.72} &
  \multicolumn{1}{c|}{16.19} &
  \multicolumn{1}{c|}{12.59} &
  \multicolumn{1}{c|}{15.73} &
  \multicolumn{1}{c|}{17.10} &
  \multicolumn{1}{c|}{20.14} &
  \multicolumn{1}{c|}{37.00} \\ \hline
\multicolumn{10}{|l|}{\cellcolor{black!10}\textit{Supervised AWEs trained on English}}                                                                                     \\ \hline
\multicolumn{1}{|l|}{MFCC} &
  \multicolumn{1}{c|}{3.83} &
  \multicolumn{1}{c|}{4.57} &
  \multicolumn{1}{c|}{10.77} &
  \multicolumn{1}{c|}{15.32} &
  \multicolumn{1}{c|}{9.16} &
  \multicolumn{1}{c|}{13.06} &
  \multicolumn{1}{c|}{11.49} &
  \multicolumn{1}{c|}{16.56} &
  \multicolumn{1}{c|}{38.15} \\ \hline
\multicolumn{1}{|l|}{wav2vec} &
  \multicolumn{1}{c|}{1.38} &
  \multicolumn{1}{c|}{1.14} &
  \multicolumn{1}{c|}{6.59} &
  \multicolumn{1}{c|}{9.44} &
  \multicolumn{1}{c|}{4.98} &
  \multicolumn{1}{c|}{7.32} &
  \multicolumn{1}{c|}{7.32} &
  \multicolumn{1}{c|}{10.12} &
  \multicolumn{1}{c|}{34.80} \\ \hline
\multicolumn{1}{|l|}{MCPC} &
  \multicolumn{1}{c|}{2.49} &
  \multicolumn{1}{c|}{2.51} &
  \multicolumn{1}{c|}{8.13} &
  \multicolumn{1}{c|}{12.23} &
  \multicolumn{1}{c|}{6.66} &
  \multicolumn{1}{c|}{10.68} &
  \multicolumn{1}{c|}{9.89} &
  \multicolumn{1}{c|}{13.99} &
  \multicolumn{1}{c|}{39.20} \\ \hline
\multicolumn{1}{|l|}{XLSR-53} &
  \multicolumn{1}{c|}{\textbf{0.93}} &
  \multicolumn{1}{c|}{\textbf{0.79}} &
  \multicolumn{1}{c|}{\textbf{4.12}} &
  \multicolumn{1}{c|}{\textbf{5.71}} &
  \multicolumn{1}{c|}{\textbf{1.92}} &
  \multicolumn{1}{c|}{\textbf{2.83}} &
  \multicolumn{1}{c|}{\textbf{5.05}} &
  \multicolumn{1}{c|}{\textbf{6.21}} &
  \textbf{31.76} \\ \hline
\end{tabular}
\caption{ABX error rates (\%) within and across speakers for each language}\label{tab:abx}
\end{table*}

\begin{table}
\centering
\begin{tabular}{cccccc|}
\cline{2-6}
\multicolumn{1}{c|}{}        & \multicolumn{1}{c|}{en} & \multicolumn{1}{c|}{fr} & \multicolumn{1}{c|}{de} & \multicolumn{1}{c|}{es} & ar \\ \hline
\multicolumn{6}{|l|}{\cellcolor{black!10} \textit{Self-Supervised AWEs for each language}}                                                                                \\ \hline
\multicolumn{1}{|l|}{MFCC}    & \multicolumn{1}{l|}{47.3}   & \multicolumn{1}{l|}{48.4}   & \multicolumn{1}{l|}{45.0}   & \multicolumn{1}{l|}{54.3}   &30.9    \\ \hline
\multicolumn{1}{|l|}{wav2vec} & \multicolumn{1}{l|}{66.1}   & \multicolumn{1}{l|}{59.9}   & \multicolumn{1}{l|}{59.5}   & \multicolumn{1}{l|}{67.5}   &35.9   \\ \hline
\multicolumn{1}{|l|}{MCPC}    & \multicolumn{1}{l|}{57.2}   & \multicolumn{1}{l|}{53.6}   & \multicolumn{1}{l|}{53.9}   & \multicolumn{1}{l|}{60.4}   &33.5   \\ \hline
\multicolumn{1}{|l|}{XLSR-53} & \multicolumn{1}{l|}{52.0}   & \multicolumn{1}{l|}{52.2}   & \multicolumn{1}{l|}{54.9}   & \multicolumn{1}{l|}{55.0}   &31.0   \\ \hline
\multicolumn{6}{|l|}{\cellcolor{black!10}\textit{Supervised AWEs trained on English}}                                                                                    \\ \hline
\multicolumn{1}{|l|}{MFCC}    & \multicolumn{1}{l|}{68.5}   & \multicolumn{1}{l|}{52.7}   & \multicolumn{1}{l|}{56.7}   & \multicolumn{1}{l|}{61.1}   & 33.4   \\ \hline
\multicolumn{1}{|l|}{wav2vec} & \multicolumn{1}{l|}{82.3}   & \multicolumn{1}{l|}{64.8}   & \multicolumn{1}{l|}{69.3}   & \multicolumn{1}{l|}{71.1}   &38.8    \\ \hline
\multicolumn{1}{|l|}{MCPC}    & \multicolumn{1}{l|}{74.5}   & \multicolumn{1}{l|}{56.4}   & \multicolumn{1}{l|}{62.7}   & \multicolumn{1}{l|}{66.2}   & 35.6   \\ \hline
\multicolumn{1}{|l|}{XLSR-53} & \multicolumn{1}{l|}{\textbf{84.3}}   & \multicolumn{1}{l|}{\textbf{69.1}}   & \multicolumn{1}{l|}{\textbf{78.1}}   & \multicolumn{1}{l|}{\textbf{75.8}}   & \textbf{41.8}    \\ \hline
\end{tabular}
\captionof{table}{K-Means Clustering Accuracy (\%)}\label{tab:cluster}
\end{table}

\subsection{Evaluation Scheme}

We constructed Minimal-Pair ABX tasks, as described in \citet{schatz2013evaluating}. ABX tasks are typically used to measure phoneme discrimination in zero-resource settings, and they consist of two segments, A and B, that differ by a minimal contrast (e.g. one phoneme difference), and a third segment X that matches either A or B. A distance measure such as DTW or cosine is used to find the closest match. We used two variants of this task: within-speaker ABX, where all three words are spoken by the same speaker, and cross-speaker ABX, where X is spoken by a different speaker. 
We automatically extracted the words from each test set; we selected A and B by finding word pairs that have the same length\footnote{Since automatic word alignments tend to be inaccurate around the boundaries, we only used words that have at least five characters. } and Levenshtein edit distance of 1 or 2, which roughly corresponds to a difference of one or two phonemes most of the time. For Arabic, the dataset did not have speaker ids, so all three words could be from different speakers. In addition, due to the lower quality of the sound recordings and the presence of noise in this dataset, the word alignment quality is much lower than the other languages, so the automatic process resulted in many invalid segments. To have a more reliable test set for Arabic, we manually checked the validity of the extracted words and kept 954 validated word pairs for evaluation.
\par
We also used clustering for complementary evaluation. We clustered the embeddings using K-Means with K being the number of unique words in the test set. We calculated the accuracy of clustering as the percentage of words that match their cluster label, which is the word id of the majority of segments in each cluster. This allows us to measure if the embeddings of the same words are similar enough to be clustered together.

\section{Results}

Table \ref{tab:abx} shows ABX error rates using the input features directly (with DTW as distance metric), self-supervised AWEs trained on each language, and supervised AWEs trained on English. Cosine similarity is the metric used in the latter two settings.  Confirming previous results \cite{riviere2020unsupervised}, we do observe that pre-trained acoustic features like Modified CPC and Wav2Vec2, which are trained exclusively on English unlabeled speech, transfer well across languages. These pre-trained features consistently outperformed MFCC features for all languages, particularly in cross-speaker evaluation. Unsurprisingly, the English language has the best ABX scores overall simply because the pre-trained features used are all trained on English. 
The results for self-supervised AWE models are mixed, but generally they are in the same range as DTW performance, which also conforms with previously published results \cite{holzenberger2018learning}.  

With supervised training, we see significant reduction in errors rates for all languages. The lowest error rates are achieved on the English test set, as expected. More notably, the largest reduction in error rates is achieved with the XLSR features. It is also interesting to note that XLSR features were not impressive in the self-supervised setting compared with other features; Wav2Vec2 and MCPC, which were trained on English only, gave better results in the self-supervised framework for all test languages. The advantage of using these cross-lingual features was only evident in the supervised and transfer learning setting, where they consistently outperformed all other features. For Arabic, the error rates are higher overall due to the nature of the dataset, but we still observe the lowest error rate in the transfer learning setting. 

Finally, we see in table \ref{tab:cluster} that the clustering accuracy results are consistent with the ABX results, where supervised models trained on English consistently gave higher accuracy compared with self-supervised models trained on the target languages.

\section{Conclusions}
Our results demonstrate the superior effectiveness of zero-shot transfer learning of acoustic word embeddings compared with self-supervised training in the target languages. This is particularly useful for low-resource languages for which data may not be available for supervised or self-supervised training. The mechanism of this transfer is mainly through the reduction in speaker variability which is far easier to achieve via supervised training. In addition, supervised training makes the most out of pre-trained features, where we see further reduction in error rates that far exceed the reduction observed in self-supervised settings. The presence of noise naturally results in larger error rates; further investigations are needed to demonstrate the transferability of noise robustness in a similar manner. 

\section*{Acknowledgement}

This work was supported by grant no. 31T139 at United Arab Emirates University and partially funded under UAEU-ZU Joint Research Grant G00003715 (Fund No.: 12T034) through Emirates Center for Mobility Research.

\bibliography{anthology,custom}
\bibliographystyle{acl_natbib}

\appendix

\section{Appendix}
\label{sec:appendix}
\subsection{Dataset Details}

\begin{table}[h]
\centering
\begin{tabular}{|l|c|c|}
\hline
                       Dataset & test   & dev    \\ \hline
English    & 52,576  & 54,402  \\ \hline
French  & 90,958  & 83,560  \\ \hline
German  & 121,713 & 122,903 \\ \hline
Spanish & 88,417  & 87,417  \\ \hline
Arabic            & 62,745  & 57,532  \\ \hline
\end{tabular}
\captionof{table}{Total number of words in each dataset}\label{tab:words}
\end{table}
\subsection{Model Architecture \& Hyper-Parameters}

The architecture described in section \ref{sec:model} was modeled after other acoustic word embedding models \cite{chung2016audio, chung2018speech2vec, holzenberger2018learning} with slight variations in details. We found that this particular configuration worked best across different acoustic features, whereas other choices gave mixed results. For example, using GRUs instead of LSTMs worked well with pre-trained features but was worse for MFCCs. The decoding process described in \citet{holzenberger2018learning}, where positional encodings are used instead of previous outputs also resulted in inferior performance. We also found that using teacher forcing instead of the model's previous output as input to the decoder hurt the performance. Finally, using two layers was crucial to get results in line with DTW perforamnce for most self-supervised models. The only exception is the self-supervised model with XLSR features which resulted in unstable training with 2 layers. We found it to work much better with a single layer network and slightly larger embedding size. Generally, larger embeddings sizes improved performance to some extent, but the improvements were smaller beyond the values that we have chosen; furthermore, using smaller sizes is more advantageous in terms of computational efficiency. We did not perform any hyper-parameter tuning for the target languages since we are working within the premise of low-resource settings where validation data may not be available.  

Table \ref{tab:params} shows the number of parameters for each model. Since the decoder is only used for training and can be discarded after that, we only show the number of encoder parameters.

\begin{table}[h]
    \centering
    \begin{tabular}{|l|c| c|c|}
    \hline
        Model &  input & hidden & no.of parameters\\
        \hline
        \multicolumn{4}{|l|}{\cellcolor{black!10} Self-Supervised} \\
        \hline
        MFCC & 39 & 100 &  354,400\\
        \hline
        Wav2Vec & 768 & 100 & 937,600\\
        \hline
        MCPC & 256 & 100 & 528,000 \\
        \hline
         XLSR-53 & 1024 & 250 &  1,411,200\\
         \hline
         \multicolumn{4}{|l|}{\cellcolor{black!10} Supervised} \\
         \hline
        MFCC & 39 & 100 &  354,400\\
        \hline
        Wav2Vec & 768 & 100 & 937,600\\
        \hline
        MCPC & 256 & 100 & 528,000 \\
        \hline
        XLSR-53 & 1024 & 100 &  1,142,400\\
        \hline
    \end{tabular}
    \caption{Input size, hidden layer size, and total number of encoder parameters for each model.}
    \label{tab:params}
\end{table}

\subsection{Training Details}

The supervised models were trained with NLL loss, and the training targets are sequences of phonemes obtained using the Phonemizer package \footnote{https://github.com/bootphon/phonemizer} \cite{Bernard2021}. This choice seemed more sensible at first, but we found that using sequences of characters instead of phonemes worked equally well. 

The model was implemented using PyTorch and trained on NVIDIA K80 GPU as provided in AWS p2.xlarge instances. For optimization, we found that adam optimizer worked for all features except MFCCs, for which SGD with cyclical or step learning rate schedule was more stable.


Table \ref{tab:words} shows the number of words in each dataset. The word alignments were obtained via force alignment using The Montreal Forced Aligner\footnote{https://github.com/MontrealCorpusTools/Montreal-Forced-Aligner} \cite{mcauliffe2017montreal} for English, German, French, and Spanish. The Montreal aligner uses an ASR engine, and since these datasets are relatively clean, the alignments are generally accurate. For Arabic, the best option was the aeneas toolkit\footnote{www.readbeyond.it/aeneas}, which relies on a TTS engine to align the synthesized words with the actual audio segments. We used Amazon Polly TTS for higher quality, but overall the alignments were not as accurate as the other datasets, which we believe is due to the low quality of the recordings, presence of noise, and high variability in accents. The low clustering accuracy could be partially attributed to the inaccurate labeling of the segments as a result of this. For ABX evaluation on the Arabic set, we manually filtered the segments that had somewhat accurate boundaries; the chosen pairs still contained high level of noise conditions, such as background music and interfering speech.

\end{document}